\documentclass[letterpaper, 10 pt, conference]{ieeeconf}  

\IEEEoverridecommandlockouts                              

\usepackage{hyperref}
\usepackage{graphicx}		
\usepackage{wrapfig}
\usepackage[format=plain,font=footnotesize,labelfont=bf,labelsep=period]{caption}
\usepackage{sidecap} 
\usepackage[export]{adjustbox}
\usepackage{subcaption}

\usepackage{amsmath}
\usepackage{amssymb}
\usepackage{bbm}
\usepackage{dsfont}
\usepackage{mathtools}
\usepackage{algorithm}
\usepackage{algpseudocode}

\usepackage{pdfsync}
\usepackage[normalem]{ulem}
\usepackage{paralist}	
\usepackage[space]{grffile} 
\usepackage[noabbrev, capitalise]{cleveref}
\usepackage{color}
\usepackage{dirtytalk}

\usepackage{amsthm}
\usepackage{balance}

\newtheorem{theorem}{Theorem}

\theoremstyle{definition}
\newtheorem{definition}{Definition}
\theoremstyle{remark}

\theoremstyle{definition}
\newtheorem{assumption}{Assumption}
\theoremstyle{definition}

\newcommand{\R}{\mathbb{R}}

\newcommand{\C}{\mathcal{C}}

\definecolor{darkblue}{RGB}{0,0,102}
\definecolor{lightblue}{RGB}{77,77,148}

\definecolor{gold}{RGB}{234, 170, 0}
\definecolor{metallic_gold}{RGB}{139, 111, 78}

\newcommand{\mb}[1]{\mathbf{ #1 }}

\DeclareMathOperator*{\argmin}{argmin}

\newcommand{\derp}[2]{\frac{\partial #1 }{\partial #2 }}
\newcommand{\x}{\mathbf{x}}
\newcommand{\lmat}{\begin{bmatrix}}
\newcommand{\rmat}{\end{bmatrix}}

\title{\LARGE \bf
Self-Supervised Online Learning for Safety-Critical \\ Control using Stereo Vision}

\author{Ryan K. Cosner$^*$, Ivan D.  Jimenez Rodriguez$^*$, Tamas G. Molnar, Wyatt Ubellacker, \\ Yisong Yue, Aaron D. Ames, and Katherine L. Bouman
\thanks{$^*$ These authors contributed equally. }
\thanks{This research is supported in part by the National Science Foundation CPS Award \#1932091, Dow (\#227027AT), BP p.l.c., AeroVironment.}
\thanks{
Authors are with the Department of Control and Dynamical Systems, the Department of Mechanical and Civil Engineering, and with the  Computing and Mathematical Sciences, California Institute of Technology, Pasadena, CA 91125, USA, {\tt\small \{rkcosner, ivan.jimenez, tmolnar, wubellac, yyue, ames, klbouman\}@caltech.edu}.}
}

\begin{document}
\maketitle
\thispagestyle{empty}
\pagestyle{empty}

\begin{abstract}
With the increasing prevalence of complex vision-based sensing methods for use in obstacle identification and state estimation, characterizing environment-dependent measurement errors has become a difficult and essential part of modern robotics.
This paper presents a self-supervised learning approach to safety-critical control.  In particular, the uncertainty associated with stereo vision is estimated, and adapted online to new visual environments, wherein this estimate is leveraged in a safety-critical controller in a robust fashion. 
    To this end, we propose an algorithm that exploits the structure of stereo-vision to learn an uncertainty estimate without the need for ground-truth data. We then robustify existing Control Barrier Function-based controllers to provide safety in the presence of this uncertainty estimate. We demonstrate the efficacy of our method on a quadrupedal robot in a variety of environments. When not using our method safety is violated. With offline training alone we observe the robot is safe, but overly-conservative. With our online method the quadruped remains safe and conservatism is reduced. 
\end{abstract}


\section{INTRODUCTION}

Accounting for vision-based uncertainty is particularly important for modern safety-critical robotic applications such as autonomous vehicles, health care, and manufacturing \cite{knight_safety_nodate}. Such safety-critical systems require controllers that provide robust safety in the presence uncertainty.
\textit{Control Barrier Functions (CBFs)} \cite{ames_control_2017, emam2019robust} 
are a popular tool which can be used to guarantee safety through the satisfaction of a Lyapunov-like inequality. CBFs have been used to achieve several safety-critical robotic tasks such as obstacle avoidance \cite{singletary_comparative_2020}, multi-agent navigation \cite{glotfelter_nonsmooth_2017}, and safe walking \cite{csomay-shanklin_episodic_2021}.
However, standard CBF theory requires accurate state estimation. This motivates the need for a method that provides safety in the presence of noisy sensor measurements.

Computer vision has become an important tool in robotics for sensing environments and identifying obstacles. It is often an integral component of robotics applications such as simultaneous localization and mapping (SLAM \cite{cadena_past_2016}).  Despite the utility and ubiquity of computer vision, using vision sensors to achieve robust safety is difficult due to the complex environment-dependent error that they generate. 
For example, error patterns are highly correlated with the textures and appearance of a scene.
Supervised methods can identify and model error as it affects the CBF \cite{dean_robust_2019, taylor_learning_nodate, csomay-shanklin_episodic_2021, magnusAdaptiveRL, choi2020reinforcement}; however, 
supervised approaches require ground-truth training data that may be difficult or impossible to obtain. Additionally,
such frameworks often inaccurately estimate errors in regions of the state space that were under-sampled in the training set;
training on indoor environments or synthetic data with easy-to-access ground-truth data often does not translate well to outdoor environments.



 \begin{figure}
     \centering
     \includegraphics[width=0.99\linewidth]{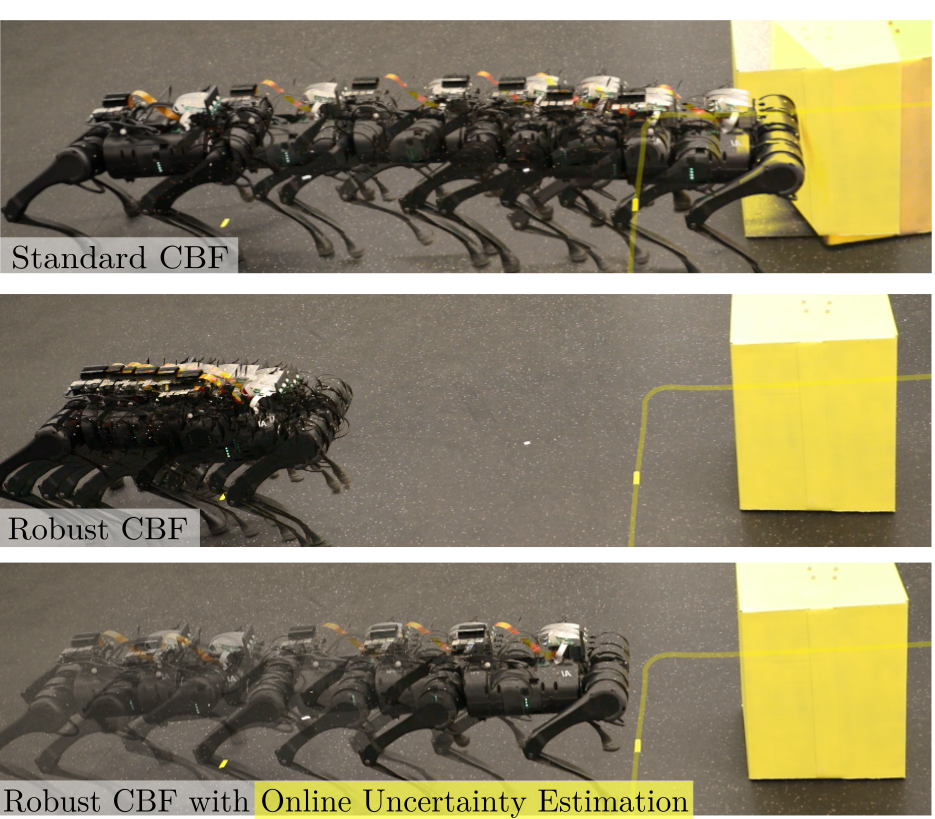}     \caption{These space-time images display our quadrupedal robot throughout the course of an experiment. The robot is considered safe if it remains left of the yellow line. The standard control barrier function (CBF) condition fails to keep the robot safe due to errors in stereo vision; the robust CBF condition keeps the robot safe, but is conservative; and our proposed method, Robust CBFs with Online Uncertainty Estimation, keeps the robot safe without remaining overly conservative. }
     \label{fig:high_level}
     \vspace{-1cm}
 \end{figure}
In this paper, we focus on two main challenges related to automatic vision-based safety critical control: (i)  accounting for the effect of perception uncertainty on the controller, 
and (ii) estimating that uncertainty while operating in novel environments.
We tackle these issues in the context of stereo vision-based obstacle avoidance on a quadrupedal robot. 

Namely, our goal is to avoid obstacles seen in stereo cameras mounted on the robot,
while accounting for the uncertainty of vision-based measurements.
To do that we use a multibaseline stereo vision system on the robot to record stereo images.
We then determine the position of the objects associated with each image pixel, and subsequently infer the uncertainty of these positions through a self-supervised error estimation algorithm that frames the problem as online learning \cite{hazan2019introduction}. Online learning frameworks have shown success in a variety of robotic applications \cite{ tomlinOnlineReachability, sofman2006improving, onlineRibeiro}
Finally, we use the position and uncertainty estimates for safety critical control, wherein we achieve robust safety via CBFs. A visualization of this method can be found in Fig.~\ref{fig:method}.

The contributions of this work are three-fold. First, we present
and evaluate an online,
self-supervised method for
characterizing the uncertainty
of disparity errors generated by stereo vision algorithms
in novel environments (Section~\ref{sec:background}).
Second, we develop a robustified CBF-based control method which utilizes this error estimate for obstacle avoidance 
(Section~\ref{sec:cbf}). And third, we demonstrate the proposed methods of error estimation and obstacle avoidance on a quadrupedal robot operating in real time (Section~\ref{sec:robot}). 


\section{STEREO VISION UNCERTAINTY QUANTIFICATION}
\label{sec:background}


We begin by revisiting stereo vision-based depth estimation.
We then propose an approach for learning the uncertainty of a black box stereo-matching algorithm. 
The proposed self-supervised learning approach can be trained online and takes advantage of geometric structure in stereo disparity maps so as not to require ground truth data. 

\subsection{Background in Stereo Vision}

\begin{figure}
    \centering
    \vspace{0.3cm}
    \includegraphics[width=0.99\linewidth]{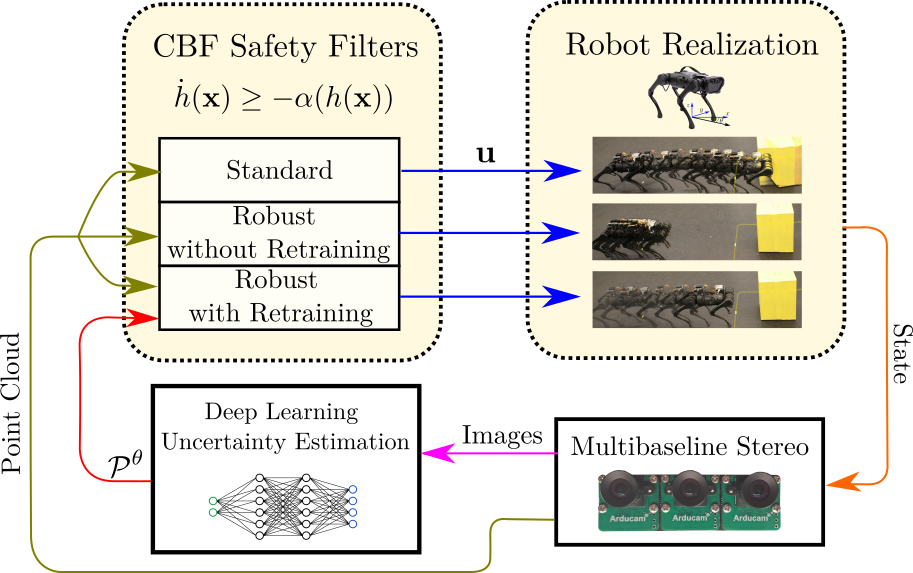}
    \caption{
    The overarching structure of our approach. It begins on the bottom right by capturing three time-synchronized images that are then fed into an uncertainty estimation pipeline and also used to generate a 3D point cloud. There are three possible CBF filters that result in the three robot realizations shown. From top to bottom, the standard filter only takes into account the noisy point-cloud in avoiding obstacles. The robust CBF safety filters use the estimate of the uncertainty $\mathcal{P}$ to compensate for noise in the point cloud.  Finally, the ``Robust with Retraining" filter refines the model of uncertainty to the current environment in real-time. 
    }
    \label{fig:method}
    \vspace{-1cm}
\end{figure}

Stereo vision is a popular tool for determining depth from images. 
These methods compute a \textit{disparity}: the shift observed in an object's projection onto two camera planes.
Using a geometric understanding of the camera setup, pixel-based disparity maps can be converted to depth maps. Errors in the final depth-map result from a combination of pixel-mismatch in disparity estimation and error in the camera parameters used to convert from disparity to depth. The errors in the intrinsic and extrinsic parameters of the camera are usually small and their effect on the resulting depth distribution is easy to compute. On the other hand, pixel matching errors are much larger and are the result of a much more complicated stereo matching procedure whose effect on the resulting disparity is difficult to quantify and environment-dependent.

For standard stereo vision we adopt the model of \cite{perrollaz2010probabilistic} for two cameras (left and right) and assume that they are perfectly rectified, vertically aligned and evenly spaced with known distance $b\in \mathbb{R}_{>0}$ between each camera. 
Pixel coordinates within an image are given by the tuple $p \triangleq (u,v) \in K$, where $K \triangleq \{0, \dots, W \} \times \{ 0, \dots, H \}$ for image width  $W \in \mathbb{N}_{>0} $ and image height $H \in \mathbb{N}_{>0}$. 

Stereo algorithms such as Block Matching, Semi-Global Block Matching, and Efficient Large-Scale Stereo \cite{geiger2010efficient} compute disparities by determining the discrete pixel distance between matching regions of two images. 
Since the disparity represents a shift between pixels of two images, the measured disparity $\widehat d$ must be a finite integer value. 
Assuming that the true disparity $d$ is a finite integer implies that the error $e\triangleq \widehat d - d $ must also be a finite integer.  
Prior work has been done to interpolate disparities for non-integer subpixel accuracy \cite{szeliski2002symmetric}; however, we restrict our attention to integer disparity values to highlight the error in pixel-matching.

\subsection{Self-supervised Error Estimation}

\label{sec:self-sup}





\renewcommand{\baselinestretch}{0.85}
To learn the error in disparity, we introduce a three-camera multibaseline stereo system which produces multiple disparity maps that are related through simple functions; deviations from the ideal relationship indicate error in the estimated disparities. 
By analyzing the correlation of image appearance with these errors, a function that estimates disparity error from appearance is learned and used to specify state error-bounds in real-time for use in a robustified CBF.



We introduce a three-element camera system, whose central camera is assumed to be perfectly rectified and vertically aligned with the other two cameras as shown in Fig.~\ref{fig:method}. This third camera is placed between the left and right cameras such that it has a baseline of $b/2$ with both. The three cameras produce a time-synchronized grayscale image triple $(I_1, I_2, I_3)$ where $I_i\in \mathbb{N}^{W \times H}$ for $i \in {1,2,3}$ and  $1,2,3$ correspond with left, center, and right, respectively. 
The disparity between any image pair $(I_i, I_j)$ for $i<j$ is obtained using the stereo-vision algorithm $\mathcal{D}: \mathbb{N}^{W\times H} \times \mathbb{N}^{W\times H} \to \Gamma^{W\times H}, $ so that $\widehat{d}_{i,j} = \mathcal{D}(I_i, I_j)$. Here, $\Gamma \subset \mathbb{N}_{\geq 0}$  is the set of possible disparity values.

Given the measurement $\widehat{d}_{i,j}$, the error appears as $\widehat d_{i,j} = d_{i,j} + e_{i,j}$ with error distribution $e_{i,j} \sim \mathcal{P}(I_i, I_j)$ and ground truth disparity $d_{i,j} \in \Gamma^{W \times H }$.
We model this error as a discrete random variable with probability $\mathcal{P}(I_i, I_j)$ on $\Gamma^{W \times H}$. This model of disparity errors contrasts sharply with other common error models, such as punctual observation, uniform observation, and Gaussian observation \cite{perrollaz2010probabilistic}, in that it accounts for the discrete nature of stereo-pixel matching algorithms. 
If ground-truth knowledge of $d_{i,j} $ is obtainable, then supervised learning methods can be implemented to directly estimate this error term. However, it is often the case that ground-truth knowledge is unavailable; particularly when a domain transfer must occur during operation. Thus we seek a general method to estimate $e_{i,j}$ for any black-box disparity algorithm without the need for ground-truth data.

We leverage the known geometric relationships between the three cameras to learn a mapping between image appearance and disparity error distribution that can adapt during operation in new environments.
Given a multibaseline stereo system, if one ignores occlusions, it is possible to completely reconstruct each disparity map from the other two maps. 
The relationship to reconstruct $\widehat d_{1,3} $ from $\widehat d_{1,2}$ and $\widehat d_{2,3}$ is shown in Algorithm~\ref{alg:disparity_sum};
we denote this reconstruction as $\overline d_{1,3} \triangleq \widehat {d}_{1,2} \oplus \widehat {d}_{2,3} $.

\begin{figure}
    \centering
    \vspace{0.25cm}
    \includegraphics[width=0.99\linewidth]{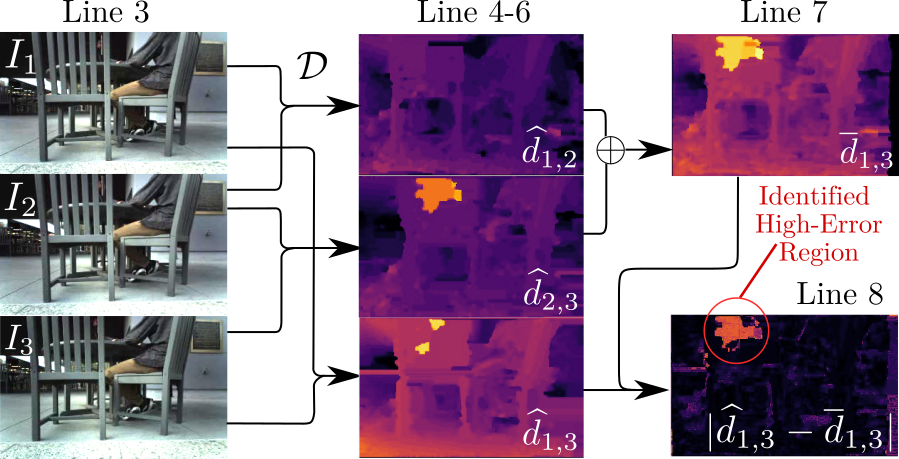}
    \caption{
    Lines 3-8 of \cref{alg:unc_est} illustrated from left to right. Starting from three time-synchronized images three pairwise disparities are computed as shown in the middle column. Two of these disparities are used to build a reconstruction of the third disparity shown in the top right which can then be used to estimate the pixel-wise error of the stereo algorithm shown in the bottom right image. These steps of the algorithm correctly identify that the back of the closest chair is a high-error region without using ground truth information. This information is used to learn a correspondence between visual features and error distributions.}
    \label{fig:learning_method}
    \vspace{-1cm}
\end{figure}

\begin{algorithm}
    \caption{Disparity Reconstruction:
    $\overline{d}_{1,3} = \widehat d_{1,2} \oplus \widehat d_{2,3}$}
    \label{alg:disparity_sum}
    \begin{algorithmic}[1]
    \State $\overline d_{1,3} \leftarrow \textbf{0}_{H \times W}$
    \For{$v \in [1, ..., H] $}%
        \For{$u \in [1, ..., W]$ }
                \State $\widehat{u} \leftarrow  n + \widehat d_{1,2}(u,v)$
                \State $\overline{d}_{1,3}(u,v)\leftarrow \widehat{d}_{1,2}(u,v) + \widehat{d}_{2,3}(u,\widehat{v})$
        \EndFor
    \EndFor
    \end{algorithmic}
    \vspace{-2pt}
\end{algorithm}%
We use the reconstructed disparity $\overline d_{1,3}$ to learn the parameters $\theta$ of a function $\mathcal{P}^{\theta}$ that approximates error distribution $\mathcal{P}$ (refer to Algorithm \ref{alg:unc_est}). 
Since this method does not require ground truth information, Algorithm \ref{alg:unc_est} can be run online during operation to adapt $\mathcal{P}^{\theta}$ to new visual environments. Recall that the disparity error, $e_{1,3}$ is discrete in nature. Therefore, the pixel-wise reconstruction error $re(p) \triangleq \Vert\widehat{d}_{1,3}^p - \overline{d}_{1,3}^p \Vert_1$ will also be discrete. For this reason, optimizing the loss $L$ reduces to a pixel-wise classification problem similar to image segmentation. Thus, as is done in image segmentation, we use pixel-wise cross entropy as the loss function $L$. This method is shown in Algorithm \ref{alg:unc_est}.
\begin{algorithm}
    \caption{Self-Supervised Stereo Error Estimation Adaptation}\label{alg:unc_est}
    \begin{algorithmic}[1]
         \State $L \leftarrow 0$
         \While{\textrm{robot is running}}
            \State $(I_1, I_2, I_3) \leftarrow \textrm{Capture Current Frame}$ \label{alg:unc_est:cap_img}
            \State $\widehat{d}_{1,2} \leftarrow \mathcal{D}(I_1, I_2)$ \label{alg:unc_est:disp1}
            \State $\widehat{d}_{2,3} \leftarrow \mathcal{D}(I_2, I_3)$ \label{alg:unc_est:disp2}
            \State $\widehat{d}_{1,3} \leftarrow \mathcal{D}(I_1, I_3)$ \label{alg:unc_est:disp3}
            \State $\overline d_{1,3} \leftarrow \widehat{d}_{1,2} \oplus \widehat{d}_{2,3}$
            \State $re(p) \leftarrow  \left\vert \widehat d_{1,3}^p - \overline{d}_{1,3}^p \right \vert$ \label{alg:unc_est:re_def}
            \State  $L \leftarrow - \frac{1}{H \times W} \sum_p \mathbb{E}_{\mathds{1}(re(p))}[\log \mathcal{P}^\theta(I_i, I_k)]$  \label{alg:unc_est:loss}
            \State $\theta_{t+1} \leftarrow \theta_t - \eta \frac{\partial L }{\partial \theta}$
            \EndWhile
        \end{algorithmic}
\end{algorithm}
In \cref{alg:unc_est:re_def}, for each pixel $p$ of the disparity $\widehat d_{1,3}$ the corresponding reconstruction error is computed. The loss function in \cref{alg:unc_est:loss} is then equivalent to the expected negative log likelihood of each pixel under the proposed model $\mathcal{P}^{\theta}$. An example visualization of lines $3-8$ can be found in Fig.~\ref{fig:learning_method}.
\noindent Although this algorithm focuses on the reconstructed disparity $\overline{d}_{1,3}$, it can be easily extended to similar reconstructions of $d_{1,2}$ and $d_{2,3}$.

Supervised methods have been used in the past to estimate uncertainty in robotic applications by computing the covariance of state estimates \cite{liu_deep_2018}.
Our approach differs in that we do not require ground truth and we take advantage of the discrete structure of images to learn a discrete, rather than a Gaussian, distribution, which is better suited to the disparity measurements of stereo vision.

\section{SAFE VISION-BASED CONTROL}
\label{sec:cbf}

In this section we review Control Barrier Functions (CBFs) \cite{ames_control_2014} as a tool for guaranteeing the safety of dynamical systems. 
We then propose CBFs that rely on the position of pixels provided by stereoscopic sensing.
Finally, we incorporate the proposed self-supervised error estimates of Section~\ref{sec:background} to enforce robust safety.

\subsection{Control Barrier Functions}


First we give a brief introduction to CBFs which follows our description in \cite{cosner_measurement-robust_2021}, where additional technical details can be found. 
In this work we consider the safety of robotic systems with control affine dynamics 
\begin{equation}
    \dot{\x} = \mb{f}(\x) + \mb{g}(\x) \mb{u}, \quad \x \in \R^n , \;\mb{u} \in \R^m, \label{eq:dynamics}
\end{equation}
where $\mb{x}$ is the state of the system, $\mb{u}$ is the input, ${\mb{f}: \R^n \to \R^n}$ is the drift dynamics, and ${\mb{g}: \R^n \to \R^{n\times m}}$ is the input matrix. We assume that $\mb{f}$ and $\mb{g}$  are locally Lipschitz continuous.
Given a locally Lipschitz continuous state-feedback controller ${\mb{k}: \R^n \to \R^m}$,  the closed-loop dynamics are governed by: 
\begin{equation}
    \dot{\x} =
    \mb{f}(\x) + \mb{g}(\x)\mb{k}(\x). \label{eq:cl}
\end{equation}
For any initial condition ${\x(0) = \x_0 \in \R^n}$ there exists
a unique solution $\x(t)$ to \eqref{eq:cl}, which we assume to exist ${\forall t \in [0, \infty)}$.


The notion of safety is formalized by defining a \textit{safe set} $\mathcal{C} \subset \R^n $ 
in the state space that the system must remain within.
In particular, consider the set $\mathcal{C}$ as the $0$-superlevel set of a continuously differentiable function $h : \R^n \to \R$: 
\begin{equation}\label{eq:safe_set}
    \mathcal{C} \triangleq \{ \x\in \R^n \; | \; h(\x) \geq 0 \},
\end{equation}
where 
$h(\x) = 0 \implies \frac{\partial h }{\partial \x}(\x)  \neq 0$ and $\mathcal{C}$ is non-empty and has no isolated points.
Safety is defined as the \textit{forward invariance} of $\mathcal{C}$, i.e., if $\mb{x}_0\in \mathcal{C}$, then $\mb{x}(t) \in \mathcal{C}$ for all $t \geq 0$. 


To synthesize controllers that ensure safety, we use Control Barrier Functions (CBFs)  \cite{ames_control_2017} defined as follows:


\begin{definition}[\textit{Control Barrier Function (CBF)}]
    \label{def:cbf}
    Let ${\C\subset\R^n}$ be a safe set given by~(\ref{eq:safe_set}).
    The function $h$ is a \textit{Control Barrier Function} (CBF) for \eqref{eq:dynamics} on $\mathcal{C}$ if there exists $\gamma\in\mathcal{K}_{\infty,e}$\footnote{$\mathcal{K}_{\infty, e}$ denotes the set of extended class-$\mathcal{K}$ infinity functions, wherein $\gamma\in \mathcal{K}_{\infty,e}$ satisfies $\gamma: \R\to\R$ is strictly increasing, $\gamma(0) =0$, and $\lim_{r\to \infty} \gamma(r) = \infty$, $\lim_{r\to-\infty } \gamma(r) = - \infty$.} such that for all $\mb{x}\in\C$:
    \begin{equation}
    \label{eqn:cbf}
         \sup_{\mb{u}\in\R^m} \dot{h}(\mb{x},\mb{u}) \triangleq \underbrace{\derp{h}{\mb{x}}(\mb{x})\mb{f}(\mb{x})}_{L_\mb{f}h(\mb{x})}+\underbrace{\derp{h}{\mb{x}}(\mb{x})\mb{g}(\mb{x})}_{L_\mb{g}h(\mb{x})}\mb{u}\geq-\gamma(h(\mb{x})),
    \end{equation}
    where $L_\mb{f}h: \R^n \to \R$ and $L_\mb{g}h: \R^n \to \R^m $ are the Lie derivatives of $h$ with respect to $\mb{f}$ and $\mb{g}$ respectively.  
\end{definition}

A main result in \cite{ames_control_2014, xu_robustness_2015} relates CBFs
to the safety of the closed-loop system \eqref{eq:cl}
with respect to $\C$:
\begin{theorem}\label{thm:cbf_safe}
Given a safe set $\C\subset\R^n$,
if $h$ is a CBF for \eqref{eq:dynamics} on $\C$, then any locally Lipschitz continuous controller $\mb{k}:\R^n\to\R^m$
satisfying
\begin{equation}
L_\mb{f}h(\mb{x})+L_\mb{g}h(\mb{x})\mb{k}(\mb{x})\geq-\gamma(h(\mb{x})) \label{eq:cbf_constraint}
\end{equation}
for all $\mb{x}\in\C$, renders the system \eqref{eq:cl} safe w.r.t. $\C$.
\end{theorem}

Given a nominal (but not necessarily safe) locally Lipschitz continuous controller $\mb{k}_d:\R^n\to\R^m$ and a CBF $h$, the CBF-Quadratic Program \eqref{eqn:CBF-QP} \cite{ames_control_2017} is a controller that guarantees the system's safety:
\begin{align}
\label{eqn:CBF-QP}
\tag{CBF-QP}
\mb{k}(\mb{x}) =  \,\,\underset{\mb{u} \in \R^m}{\argmin}  &  \quad \frac{1}{2} \| \mb{u} -\mb{k}_d(\mb{x})\|_2^2  \\
\mathrm{s.t.} \quad & \quad L_\mb{f}h(\mb{x}) + L_\mb{g}h(\mb{x}) \mb{u} \geq - \gamma (h(\mb{x})). \nonumber
\end{align}

\subsection{Control Barrier Functions for Safe Vision-Based Control}
\label{sec:cbfs_vision}

Next we apply CBFs to achieve safe obstacle avoidance for robotic systems based on stereo vision. First we construct CBFs for safe vision-based control.
Let $\rho_{p} \in \R^3$ represent the true three-dimensional position of the portion of the scene which generated pixel $p$. Using this, we can define a CBF $h: \R^n\times\R^3 \to \R$ that relies on both the state $\x$ and three dimensional pixel position $\rho_p$. The pixel position is a geometric function of the true disparity, 
$\rho_p = T(\mb{x}, r(p,d^p_{1,3})) $ where $r: \mathbb{N}^2 \times \mathbb{N}$ is the stereo reprojection function and $T: \R^n\times \R^3 \to \R^3$ is the transformation mapping from the robot's state and relative pixel position to pixel position.

In order to relate the output of the stereoscopic sensor with safety, we make the following assumptions:

\begin{assumption}\label{assp:static_env}
    The environment is static, so the time derivative of the pixelized environment is zero: $\frac{d \rho_p}{ dt } = 0 $ for all $p \in K$. 
\end{assumption}

\begin{assumption} \label{assp:pxl_to_safety}
     Ensuring safety with respect to the true three dimensional pixel locations is sufficient to ensure safety with respect to the environment. That is, the safe set for the system is given by: \begin{align}\label{eq:safe_set_pixels}
        \mathcal{C}_K = \{ \mb{x} \in \R^n \; | \; h(\mb{x}, \rho_p) \geq 0 , \; \forall p \in K \} 
    \end{align}
    where $h : \R^n \times \R^3 \to \R $ is the CBF for pixel $p$. 
\end{assumption}

Although it is not outlined in this work, Assumption \ref{assp:static_env} can be relaxed to include moving environments by calculating $\dot h $ accordingly and estimating the motion of obstacles in the environment. Assumption \ref{assp:pxl_to_safety} simplifies the surrounding environment from infinite- to finite-dimensional by assuming that the environment is smooth between a sufficiently dense coverage of pixels. It also implies that the system only has to stay safe with respect to objects that can be seen in the cameras' field of view.\footnote{The field of view aspect of Assumption \ref{assp:pxl_to_safety} can be overcome by tracking features that leave the frame as done in Simultaneous Localization and Mapping (SLAM) algorithms \cite{cadena_past_2016}.}  

Given Assumptions~\ref{assp:static_env} and \ref{assp:pxl_to_safety} and Theorem~\ref{thm:cbf_safe}, synthesizing the control input $\mb{u}$ such that
\begin{equation}
L_\mb{f}h(\mb{x}, \rho_p)+L_\mb{g}h(\mb{x}, \rho_p)\mb{u} \geq -\gamma(h(\mb{x}, \rho_p)),
\label{eq:safety_allpixels}
\end{equation}
${\forall p \in K}$, is sufficient to guarantee safety.
Considering each pixel ${p \in K}$, however, may be computationally intractible, therefore we seek a condition with fewer required constraints.

To combine the constraints, we apply Boolean composition to each CBF $h $ to produce a single nonsmooth CBF $h_\mathrm{ns}$,
\begin{equation}
    h_\mathrm{ns}(\mb{x}) \triangleq \min_{p \in K} h (\mb{x}, \rho_p), \label{eq:hmin}
\end{equation}
and simply enforce the CBF constraint associated with the pixels whose CBFs have the smallest value \cite{glotfelter_nonsmooth_2020}. In particular, to achieve safety it is sufficient to enforce only the constraints whose indices appear in the locally-encapsulating index set:
\begin{equation} \label{eq:std_index_set}
\Lambda = \{ p  \in K : h(\mb{x}, \rho_p) \leq h_\mathrm{ns}(\mb{x}) + \delta \},
\end{equation}
for some $\delta> 0 $, as stated formally below.
  
\begin{theorem}[\cite{glotfelter_nonsmooth_2020}, Prop III.6] \label{thm:nonsmooth}
    Let $h:\R^n\times\R^3 \to \R^3$ be a locally Lipschitz function and $h_\textrm{ns}$ be as in \eqref{eq:hmin}. If there exists a locally Lipschitz extended class $\mathcal{K}$ function $\gamma$ and a measurable and locally bounded controller $\mb{k} :  \R^{n} \to \R^m$ that satisfies: 
    \begin{equation}
        \min_{p \in \Lambda} \left\{
        L_\mb{f}h(\mb{x}, \rho_p) + L_\mb{g}h(\mb{x}, \rho_p) \mb{k}(\mb{x})  \right\}
        \geq - \gamma (h_\mathrm{ns}(\mb{x})). 
        \label{eq:safety_nonsmooth}
    \end{equation}
    Then $h_\mathrm{ns}$ is a valid nonsmooth CBF and the closed loop dynamics \eqref{eq:cl} with controller $\mb{k}$ are safe with respect to $\mathcal{C}_K$. 
\end{theorem}
\noindent This theorem indicates that enforcing the CBF condition only for the ``least safe" pixel is sufficient to guarantee the safety of the system. 


\subsection{Robustness to Uncertainty} \label{sec:adding_robustness}


Error in the disparity propagates to the controller in the form of the measured 3D pixel position ${\widehat{\rho}_{p}} $.
The measured value $\widehat{\rho}_{p}$ lies in a neighborhood $\mathcal{E}_p$ of the true value $\rho_{p}$, which is characterized by the error distribution $\mathcal{P}(I_i, I_j)$. We assume that the distribution $\mathcal{P}(I_i, I_j) $ is symmetric about the measured value and define the pixel-wise uncertainty set: 
\begin{equation}
    \mathcal{E}_p \triangleq \left\{ \rho \in \mathbb{R}^3 \; \bigg | \; \begin{array}{c}
          \rho =  T(\mb{x}, r(p, \xi)), \quad \xi \in \Gamma\\
          \mathcal{P}^\theta(e_{1,3}(p) < \vert \xi - \widehat d(p) \vert; I_1, I_3) \geq  \sigma \end{array}    \right\}   
\end{equation}
where $\sigma>0$ is a parameter defining the desired uncertainty robustness. 

To achieve safety, one must determine which pixels are safety-critical given $\mathcal{E}_p$ and then enforce robust safety with respect to those pixels. The safety-critical pixels can be determined by expanding the index set $\Lambda$ using the uncertainty:
\begin{align}
    \Lambda & \subseteq \left\{  p \in K  \; \bigg | 
    h(\mb{x}, \rho_p)  \leq \max_{\rho_p \in \mathcal{E}_p}\min_{p \in K } h(\mathbf{x}, \rho_p) + \delta   \right\} .
\end{align}
This can further be expanded to an easily calculable index set $\widehat{\Lambda} \supseteq \Lambda$ by minimizing the left-hand-side of the inequality condition and using the max-min inequality \cite{boyd2004convex}:
\begin{align}
    \widehat{\Lambda} = \left\{ p \in K  \bigg\vert
    \min_{\rho_p \in \mathcal{E}_p} h(\mb{x}, \rho_p)  \leq \min_{p \in K  } \max_{\rho_p \in \mathcal{E}_p} h(\mathbf{x}, \rho_p) + \delta  \right\}.
    \label{eq:index_maxmin}
\end{align}


\noindent This expanded index set $\widehat{\Lambda}$ accounts for uncertainty and indicates which pixels are safety-critical and which constraints must be enforced to achieve safety given the pixel-wise uncertainty sets $\mathcal{E}_p$.

 
Measurement-Robust Control Barrier Functions (MR-CBFs) as outlined in \cite{dean_guaranteeing_2020} are a general method for accounting for state uncertainty in CBFs. We can use this method for each pixel $p \in \widehat \Lambda$ to ensure that the safety constraint is satisfied despite the uncertainty. The resulting constraint is:
\begin{multline}
    L_\mb{f}h(\mb{x}, \widehat\rho_p)+  L_\mb{g}h(\mb{x}, \widehat \rho_p) \mb{u} \\
    - \left( \mathfrak{L}_{L_fh} + \mathfrak{L}_{\gamma \circ h_\mathrm{ns}} + \mathfrak{L}_{L_gh} \Vert \mb{u } \Vert_2 \right) \epsilon_p \\
    \geq -  \gamma (h_\mathrm{ns}(\mb{x})), \quad \forall p \in \widehat \Lambda 
    \label{eq:safety_uncertain}
\end{multline}

\noindent where $\mathfrak{L}$ is the Lipschitz constant of the subscript and 
\begin{equation}
    \epsilon_p \geq \max_{\rho_p \in \mathcal{E}_p} \Vert \rho_p - \widehat{\rho}_p\Vert_2
\end{equation} 
is a bound on the uncertainty. Since $\Lambda \subseteq \widehat{\Lambda}$ and the MR-CBF condition implies the CBF condition \eqref{eq:cbf_constraint}, satisfying \eqref{eq:safety_uncertain} also satisfies \eqref{eq:safety_nonsmooth} providing safety of the system if $\sigma=1$ and $\mathcal{P}^\theta = \mathcal{P}$.




\section{APPLICATION: OBSTACLE AVOIDANCE ON A QUADRUPEDAL ROBOT}
\label{sec:robot}

\begin{figure*}
    \centering
    \vspace{0.25cm}
    \includegraphics[width=0.99\linewidth]{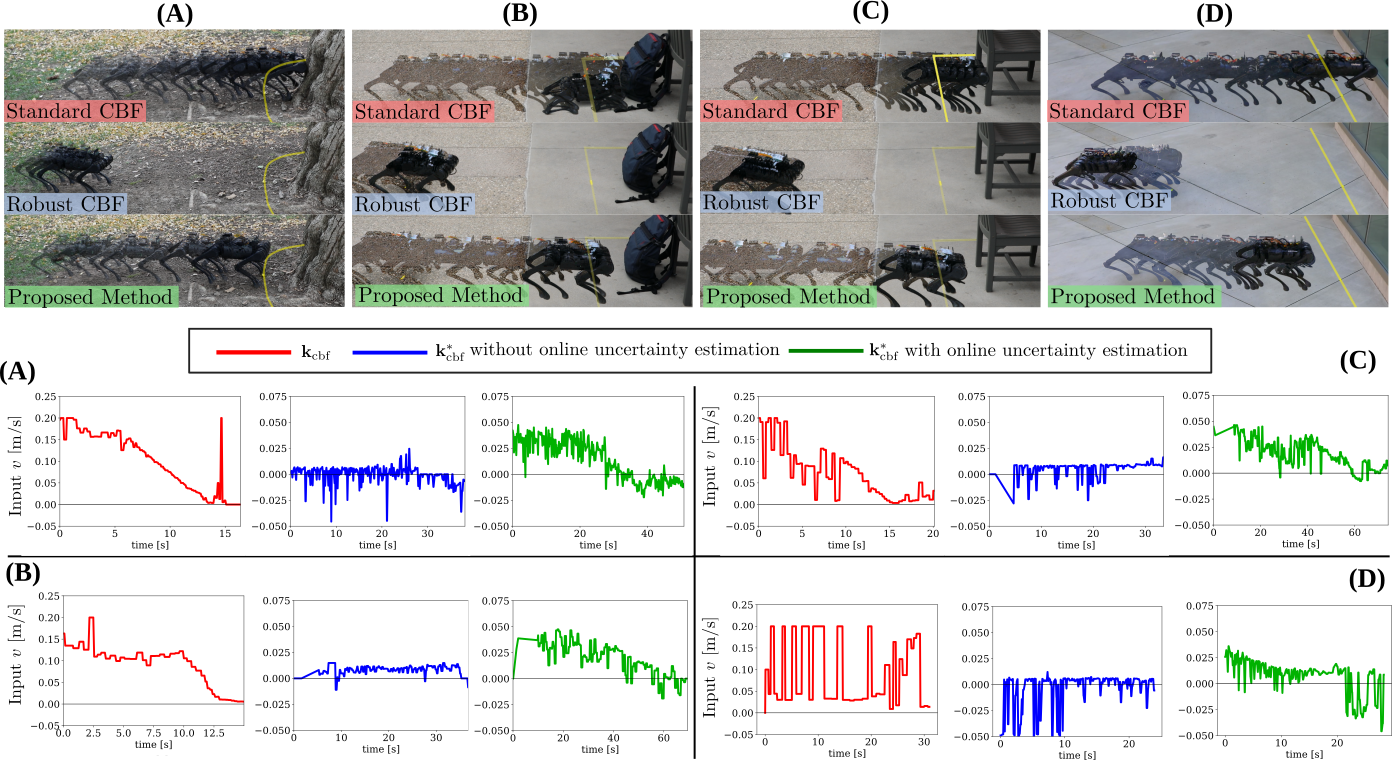}
    \caption{Demonstration of our method in a variety of environments. From left to right the goal is to maintain a safe distance from  (\textbf{A}) a tree, (\textbf{B}) a backpack, (\textbf{C}) a chair, (\textbf{D}) and a glass window. The distance to the barrier is measured and marked on the floor with a yellow tape for visualization purposes -- we emphasize this tape is not used for depth estimation.
    Notice that the barrier is assumed to be a sphere around an obstacle but in the case on the glass, this sphere degenerates into a plane.
    The quadrupedal robot is given a desired control input of 0.2 m/s.
    In all cases, a naive barrier implementation that simply takes the noisy measurements from a stereo vision system fails to keep the system safe. The robustified controller \eqref{eq:exp_controller} with a pretrained model consistently shows overly conservative behavior. 
    Finally, with online learning, the robot converges to the barrier without exhibiting conservative behavior, except for the glass environment where the robot is overly conservative and walks away from the barrier due to the perceived uncertainty.
    The (\textbf{A-D}) corresponding plots below show the control input filtered by the barrier in each of the three robustification cases.}
    \label{fig:results}
    \vspace{-0.6cm}
\end{figure*}


In this section, we evaluate our approach on a quadrupedal robotic platform. With these experiments we aim to demonstrate: 1) Our method is capable of keeping the system safe in a simple do-not-collide task, and 2) Our method can adapt online to measurement uncertainty in different environments without ground-truth data.

\subsection{Hardware System}
For the hardware experiments we designed a custom camera array with three equally spaced inexpensive CMOS, global shutter, time-synchronized Arducam cameras. An Nvidia Jetson Nano is used to capture, downsize, and greyscale the stereo images. The images are then sent to an external computer that receives the images and outputs the filtered control input at a frequency of at least 10 Hz. The robot used in this experiment is a Unitree A1 quadrupedal robot that receives inputs of velocity and angle rate, $\mb{u} = \lmat v & \omega \rmat^\intercal$. A 1 kHz Inverse Dynamics Quadratic Program (ID-QP) walking controller designed using the concepts in~\cite{buchli2009inverse}, is used to track these inputs. Stereo pixel-matching calculations were performed using Efficient LArge-scale Stereo (ELAS) \cite{geiger2010efficient}.

\subsection{Learning Method and Model}
The architecture of the model used to estimate $\mathcal{P}^{\theta}$ is a modified version of the Hierarchical Multi-Scale Attention for Semantic Segmentation introduced in \cite{tao2020hierarchical}; this model is relatively lightweight, consisting of only 196 thousand parameters (e.g., network weights). The robustness threshold used was $\sigma = 0.99$ and the online learning rate was 0.001.  We pretrain the model until convergence on a dataset of 6000 stereo image triples collected by manually moving the camera array through a variety of environments.

\subsection{Dynamics Model and Control}

In order to control the system we consider a reduced order model of the system dynamics given by the standard unicycle model. The specific form of \eqref{eq:cl} for this system is: 
\begin{equation}
    \underbrace{
    \lmat \dot{x} \\ \dot{y} \\ \dot{\theta} \rmat}_{\dot{\mb{x}}} = \underbrace{\lmat 0 \\ 0 \\ 0  \rmat }_{\mathbf{f}(\mb{x})} + \underbrace{\lmat \cos \theta & 0 \\ \sin \theta & 0 \\ 0 & 1 \rmat}_{\mathbf{g}(\mb{x})} \underbrace{ \lmat v \\ \omega \rmat}_{\mb{k}(\mb{x})}   
    \label{eq:unicycle_dyn}
\end{equation}
A formal analysis of CBFs which utilize reduced-order velocity input models is described in \cite{molnar2021modelfree}.

For this system we consider the pixel-wise CBFs, 
\begin{align} \label{eq:ball_barrier}
    h(\mb{x}, \rho_p)  & = \frac{1}{2}\left( \left\Vert  \lmat x \\ y \rmat -  \lmat \rho_{p,x} \\ \rho_{p,y} \rmat \right\Vert^2_2 - c^2\right)
\end{align}
\noindent where $\rho_{p,x}$ and $\rho_{p,y}$ indicate the global real-world $x$ and $y$ positions of pixel $p$. This function characterizes safety as remaining a planar distance $c>0$ from $\rho_{p}$. This can be thought of as buffering surfaces in the environment by a radius $c$. 



\subsection{Robustness to Uncertainty}

To illustrate the efficacy of our method we use two controllers in our experiments. A standard, unrobustified controller: 
\begin{align}
    \mb{k}_\textrm{cbf} = \argmin_{\mb{u} \in \R^2} & \quad \frac{1}{2}\Vert \mb{k}_{\textrm{des}}(\mb{x}) - \mb{u} \Vert_2^2 \label{eq:std_exp_controller}\\
    \textrm{ s.t. }    \quad & \underbrace{-\lmat 1 & 0 & 0 \rmat^\intercal r(p,\widehat{d_p})v}_{\dot{h}} \geq  -\gamma (\min_{p \in K} h(\mb{x}, \widehat{\rho_p})), \nonumber \\
    & \quad \quad \quad \quad \quad \quad  \quad \quad \quad \quad \quad \quad \quad \quad \quad  \forall p \in  \Lambda\nonumber
\end{align}

\noindent and a robustified controller: 
\begin{align}
    \mb{k}_\textrm{cbf}^* = \argmin_{\mb{u} \in \R^2} & \quad \frac{1}{2}\Vert \mb{k}_{\textrm{des}}(\mb{x}) - \mb{u} \Vert_2^2 \label{eq:exp_controller}\\
    \textrm{ s.t. } & -v \geq  \frac{-\gamma (\min_{p \in K} h(\mb{x}, \rho_p^*))}{\lmat 1 & 0 & 0 \rmat^\intercal r(p,d_p^*) },  \quad \forall p \in \widehat{ \Lambda} . \nonumber
\end{align}
where $\mb{k}_\textrm{des}: \R^m \to \R^n $ is a desired controller, $d^*$ is the maximum disparity for any $\rho_p \in \mathcal{E}_p$, and $\rho_p^*$ is pixel location associated with $d_p^*$.

Controller \eqref{eq:exp_controller} is obtained by first replacing the index set $\Lambda $ with the $\widehat{\Lambda}$. Next we note that $\lmat 1 & 0 & 0 \rmat^\intercal r(p,\widehat{d}_p) $ is  strictly positive. After dividing by this quantity, the constraint in \eqref{eq:std_exp_controller} is robustified to account for the worst-case error as is done with MR-CBFs.
Experimentally, this controller was implemented with  $\delta = 0$ and a maximum of 4000 constraints. 


\subsection{Experimental Results}
The system was run in 4 different environments (see Fig.~\ref{fig:results}). The CBF \eqref{eq:ball_barrier} was used with a safe radius of $c = 0.33$ m. The intended obstacle in the 4 different environments were  (\textbf{A}) a tree, (\textbf{B}) a backpack, (\textbf{C}) a chair, (\textbf{D}) and a glass window. A desired constant forward velocity $v = 0.2 $ m/s was used in each experiment and the robot was started approximately 1.3 m away from the obstacle. Since ground-truth measurements were unavailable, we use a yellow line on the ground to indicate the true location of the barrier. 

For each environment three different tests were performed. First, controller \eqref{eq:std_exp_controller} was used. Since this did not consider measurement uncertainty it failed to achieve safety in every environment; in all experiments the stereo vision overestimated the distance to objects at some point during the run and the quadruped ran directly into the obstacles. Second, the controller \eqref{eq:exp_controller} was used with an error estimate computed through a pretrained function  $\mathcal{P^\theta}$; this succeeded in providing safety, but was found to be overly conservative and did not allow the quadruped to approach the obstacle as desired. Third, the controller \eqref{eq:exp_controller} was used with a $\mathcal{P}^\theta$ that adapted to the environment according to Algorithm \ref{alg:unc_est}. In this case, safety of the system was generally maintained and over time the system was able to approach the boundary of the safe set. Even when small safety violations occurred, the system eventually corrected and came to rest at a safe steady-state. These results can be seen in Figure \ref{fig:results}. A video of the experiments can be found at \cite{videoVideo}.

\section{CONCLUSION AND FUTURE WORK}

We presented a framework for achieving safety of a stereo vision-based system using self-supervised online uncertainty estimation and robustified CBFs. Refining the uncertainty estimate model online was shown to achieve significantly better performance. We validated our online learning approach across several environments and successfully achieved robust safety with minimal violations and conservatism.

Future work involves providing mathematical guarantees for our method and extending this theory and application to dynamic environments.
We also note that our online uncertainty estimation method, as outlined in Algorithm \ref{alg:unc_est}, is a general method that can be coupled with other control or state estimation techniques. Finally, implementing a high-performance on-board version of our hardware system would remove the need for a remote computer.
\balance
\bibliographystyle{IEEEtran}
\bibliography{cosner_main}


\end{document}